\documentclass[11pt]{article}

\usepackage[final]{acl}

\usepackage{times}
\usepackage{latexsym}

\usepackage[T1]{fontenc}

\usepackage[utf8]{inputenc}

\usepackage{microtype}

\usepackage{inconsolata}

\usepackage{graphicx}
\usepackage{inconsolata}
\usepackage{multirow}
\usepackage{framed}
\usepackage{makecell} 
\usepackage{lipsum} 
\usepackage{enumitem}
\usepackage{listings}
\usepackage{paralist}
\usepackage{booktabs}
\usepackage{subcaption}
\usepackage{graphicx}
\usepackage{amsmath}
\usepackage{amssymb}
\usepackage{bbm}
\usepackage{pifont}
\usepackage{marginnote}
\usepackage[color=white,size=footnotesize]{todonotes}
\usepackage{verbatim}
\usepackage{tcolorbox}

\title{Generating and Refining Dynamic Evaluation Rubrics for LLM-as-a-Judge}

 \author{Zijie Wang \quad
         Eduardo Blanco\\
         University of Arizona \\
         Department of Computer Science \\
         \texttt{\{zijiewang, eduardoblanco\}@arizona.edu} 
 }

\begin{document}
\maketitle
\begin{abstract}
LLM-as-a-Judge is a scalable alternative to human evaluation, yet existing rubric-based methods rely on human-annotated data such as reference answers or expert-crafted rubrics.
We propose to automatically generate fine-grained evaluation rubrics without any human annotation.
Our training-free method generates rubrics at dataset-specific and instance-specific granularities, achieving performance competitive with existing methods across four benchmarks.
We further present a method that iteratively fine-tunes a rubric generator model via meta-judge reward signals.
The fine-tuned generator outperforms all existing baselines in both pairwise and pointwise evaluation.
Notably, a fine-tuned 14B rubric generator outperforms a much larger proprietary model at rubric generation, 
showing the effectiveness of our fine-tuning strategy.
\end{abstract}

\section{Introduction}
LLM-as-a-Judge has emerged as a promising paradigm for evaluating LLM outputs, mitigating limitations of traditional similarity-based metrics such as BLEU, ROUGE, and BERTScore~\cite{lin-2004-rouge, papineni-etal-2002-bleu, Zhang2020BERTScore}.
This approach has demonstrated success across diverse applications, including evaluation of natural language generation~\cite{zeng2024evaluating}, validation of synthetic data~\cite{leang-etal-2025-theorem}, and self-rewarding for LLM alignment~\cite{yuan2024selfrewarding}.
However, LLM judges exhibit biases toward irrelevant heuristics, such as answer length~\cite{dubois2024length} and the order of candidate responses~\cite{wang-etal-2024-large-language-models-fair}, among others.

To address these biases, previous work has explored training specialized LLM evaluators~\cite{kim2024prometheus,kim-etal-2024-prometheus,deshpande2024glider}.
However, such methods require substantial computational costs and high-quality training data.
An alternative line of research focuses on training-free methods that leverage evaluation rubrics---structured sets of criteria that specify what dimensions to assess and how to score them---to guide LLM judges toward more consistent and interpretable evaluation.
Within this direction, several studies employ fine-grained, decomposition-based rubrics. \citet{kim-etal-2025-biggen} develop a meta-evaluation benchmark with instance-specific rubrics, and several works decompose holistic evaluation into checklists or weighted sub-criteria~\cite{furuhashi-etal-2025-checklists,lee-etal-2025-checkeval,li-etal-2025-dna,li2026rubrichub}.
Other efforts curate reusable evaluation metrics across tasks, such as AutoMetric~\cite{ryan2026autometrics}.
Despite these efforts, existing approaches typically rely on human-annotated data such as expert-crafted rubric examples and reference answers, limiting their applicability to critical domains such as finance, health, and security, where expert annotations are scarce or unavailable.


%

This work focuses on improving LLM-as-a-Judge systems from the evaluation rubric perspective. 
We investigate automatically generating fine-grained rubrics without requiring any human-annotated data (e.g., reference answers, rubric examples, or annotated training data).
Specifically, we propose a two-stage approach.
First, we introduce a training-free method that directly generates evaluation rubrics at multiple granularities---from dataset-specific to instance-specific---without requiring human-annotated examples or reference answers.
Second, we investigate fine-tuning a rubric generator model via preference learning, leveraging reward signals derived from a meta-judge's evaluation on the generated rubrics.
Experiments across four LLMs and four benchmarks demonstrate that our training-free approach achieves competitive performance with several baselines and human-crafted rubrics, 
while rubrics from our fine-tuned generator further improve the LLM judge, outperforming all existing methods.

Our contributions are as follows:\footnote{Data, code and models available at \url{https://github.com/wang-zijie/generating_dynamic_rubric}}
\begin{compactitem}
  \item We present a training-free approach for generating fine-grained rubrics that yields competitive performance with existing methods.
  \item We propose a preference fine-tuning method for rubric generator models using meta-judge feedback, requiring no human-annotated data.
  \item We conduct experiments with four LLMs and four benchmarks, demonstrating state-of-the-art performance in both pairwise and pointwise evaluation settings.
  \item We provide qualitative analyses and case studies revealing how preference fine-tuning shifts rubric framing, explaining the observed performance gains.
\end{compactitem}
\section{Related Work}
\subsection{LLM-based Automatic Evaluation}
With advances in LLM fine-tuning methods, recent work has explored various learning-based approaches to improve LLM judge models.
Prometheus 2~\cite{kim-etal-2024-prometheus} is an open-source LLM evaluator trained on a human-crafted dataset
that includes predefined rubrics and reference answers~\cite{kim2024prometheus}.
GLIDER~\cite{deshpande2024glider} is another open-source LLM evaluator trained on synthetic data that employs span highlighting to provide explainable, fine-grained evaluations.
Hercule~\cite{doddapaneni-etal-2025-cross} is a multilingual LLM evaluator fine-tuned on multilingual data obtained via machine translation.
JudgeLRM~\cite{chen2025judgelrm} focuses on improving LLM judges' reasoning ability by adopting RL-based preference training with a specialized reward function.
MetaRewarding~\cite{wu-etal-2025-meta} introduces a self-improvement framework that leverages LLMs to judge their own judgments and uses that feedback to refine judging ability.

In this work, we demonstrate that a training-free approach yields competitive evaluation results against state-of-the-art LLMs and LLM judges.

\subsection{Fine-grained and Dynamic Rubrics for Automatic Evaluation}
Another line of work focuses on training-free approaches to improve LLM judges' evaluation quality, specifically by improving the quality of evaluation rubrics.
FLASK~\cite{ye2024flask} proposes a fine-grained evaluation framework that decomposes response quality into skill-specific dimensions aligned with human judgment, though it relies on a fixed, manually curated skill taxonomy.
\textsc{BiGGen Bench}~\cite{kim-etal-2025-biggen} presents a meta-evaluation benchmark with instance-specific rubrics and demonstrates that they are beneficial for LLM judges.
RubricHub~\cite{li2026rubrichub} introduces an automated rubric generation framework grounded in reference answers and predefined principles.
AutoMetric~\cite{ryan2026autometrics} proposes a task-specific evaluation framework that generates evaluation metrics with rubrics, retrieves from a curated MetricBank, and optimizes their combination via regression to maximize correlation with human judgment; 
however, this approach relies on ground truth for both generation and optimization.
\citet{li-etal-2025-dna} present an approach that decomposes evaluation into multiple criteria and aggregates scores with dynamic weights; however, their method is limited to pairwise evaluation.
CheckEval~\cite{lee-etal-2025-checkeval} adopts a checklist-based evaluation method by decomposing high-level rubrics into discrete boolean questions, but relies on human knowledge to select evaluation dimensions and sub-dimensions (e.g., dimension ``fluency'' with sub-dimensions ``grammar'' and ``readability'').
\citet{furuhashi-etal-2025-checklists} further investigate checklist-based evaluation with different checklist generation policies and observe that such methods tend to work well only in pairwise settings while showing fewer benefits for pointwise evaluation.

These approaches typically rely on human-annotated data such as reference answers or predefined evaluation dimensions.
Our method supports both pointwise and pairwise settings without any such data, making it applicable to domains where these resources are unavailable.
\begin{table*}[t]
  \centering
  \small
  \resizebox{\textwidth}{!}{%
\begin{tabular}{l r l c c c c}
\toprule
Dataset & \# Data & Eval.\ Scheme & Scale & \makecell{Ref.\\Answer} & \makecell{Human-crafted\\Rubrics} & \makecell{Instance-specific\\Rubrics} \\
\midrule
HelpSteer2~\cite{wang2024helpsteer}                          & 1,038  & Pointwise & 0--4  & \ding{55}  & \ding{51} & \ding{55}\\
\textsc{BiGGen Bench}~\cite{kim-etal-2025-biggen}            & 2,776  & Pointwise & 1--5  & \ding{51} & \ding{51} & \ding{51} \\
\addlinespace
AlpacaEval~\cite{dubois2024length}                            & 321    & Pairwise  & ---   & \ding{55}  & \ding{51} & \ding{55}  \\
MT-Bench~\cite{10.5555/3666122.3668142}                      & 941    & Pairwise  & ---   & \ding{55} & \ding{51} & \ding{55}   \\
\textsc{BiGGen Bench}~\cite{kim-etal-2025-biggen}            & 2,792  & Pairwise  & ---   & \ding{51} & \ding{51} & \ding{51} \\
\bottomrule
\end{tabular}%
}

  \caption{Overview of evaluation benchmarks. We evaluate on three pairwise and two pointwise benchmarks spanning general instruction-following, multi-turn dialogue, and specialized domains. ``Instance-specific Rubrics'' indicates whether the benchmark provides per-instance evaluation criteria; ``Human-annotated Rubrics'' indicates whether those criteria were authored by human experts.}
  \label{table:dataset}
\end{table*}

\subsection{Rubric-based Reward Modeling}
Beyond improving LLM-as-a-Judge systems, recent work has investigated leveraging rubric-based reward signals for RL fine-tuning.
Rubrics as Rewards~\cite{gunjal2026rubrics} investigates an on-policy reinforcement learning method that leverages generated rubrics as reward signals.
\citet{gupta-etal-2025-carmo} address reward hacking by dynamically generating task-specific rubrics rather than relying on static rubrics for reward modeling.
\textsc{AutoRule}~\cite{wang2025autorule} automatically extracts rules from human preference data to construct rule-based rewards for improving instruction-following ability.
However, all of these approaches rely on gold reference answers as a supervised proxy for generating or extracting rubrics.
Lastly, checklist-based rubrics have also been adopted for preference tuning. \citet{viswanathan2025checklists} propose replacing traditional reward models with dynamic, instruction-specific checklists evaluated by LLM judges.
We are the first to investigate fine-tuning a rubric generator model via reward signals from a meta-judge.


%
\section{Generating Dynamic Rubrics}

We formalize evaluating LLM-as-a-Judge systems with rubrics and describe our approach to generating fine-grained evaluation rubrics without human preference data.
We first define the two evaluation settings (Section~\ref{sec:task}) and evaluation benchmarks (Section~\ref{sec:benchmarks}),
followed by our training-free rubric generation approach (Section~\ref{sec:rubric-gen}) and the preference learning approach (Section~\ref{sec:dpo}).

\subsection{Task Formulation}
\label{sec:task}

We consider two evaluation settings commonly adopted in the LLM-as-a-Judge paradigm: \emph{pairwise} and \emph{pointwise} evaluation.

\paragraph{Pairwise Evaluation}
Given an instance input $x$ and two candidate responses $y_1, y_2$, the judge model $\mathcal{J}$ decides $\mathcal{J}(x, y_1, y_2) \in \{1, 2\}$, indicating which response is preferred.
We evaluate pairwise judges by measuring \emph{human agreement}: the percentage of instances where the judge's preference matches the majority vote of human annotators.

\paragraph{Pointwise Evaluation}
Given an instance input $x$ and a single response $y$, the judge model $\mathcal{J}$ produces a scalar score $\mathcal{J}(x, y) \in [a, b]$, 
where $[a,b]$ denotes the scoring range 
(e.g., an integer in $[1, 5]$ for holistic scoring, or a continuous value in $[0, 1]$ for criterion-level aggregation).
We measure alignment with the ground truth (i.e., human scores) using Spearman and Pearson correlation coefficients computed across all (instance, response) pairs in the dataset.
Note that since these correlation metrics are scale-invariant, the judge and human scores need not share the same range.

\paragraph{Rubric Granularity}
We distinguish two levels of rubric specificity used in this work.
A \emph{dataset-specific} rubric defines a fixed evaluation instruction applied uniformly to all instances within a benchmark or domain, capturing general quality dimensions shared across instances.
An \emph{instance-specific} rubric is generated for each individual instance, providing criteria tailored to the particular instance.

\subsection{Benchmarks}
\label{sec:benchmarks}
We evaluate on four meta-evaluation benchmarks spanning pairwise and pointwise settings (Table~\ref{table:dataset}).

\paragraph{Pointwise Benchmarks}
HelpSteer2~\cite{wang2024helpsteer} contains 1,038 instances scored on five dimensions (helpfulness, correctness, coherence, complexity, verbosity) as integers in $[0,4]$.
We use the average of helpfulness, correctness, and coherence as the ground-truth score, as complexity and verbosity lack a monotonic relationship with quality (e.g., higher verbosity is not inherently better or worse).
It provides no reference answers or instance-specific rubrics.
\textsc{BiGGen Bench}~\cite{kim-etal-2025-biggen} provides human scores ($[1,5]$) for 695 prompts across 4 models (2,776 instances), along with per-instance scoring rubrics defining what each score level means.
It provides both reference answers and instance-specific rubrics.

\paragraph{Pairwise Benchmarks}
Length-Controlled AlpacaEval~\cite{dubois2024length} contains 321 response pairs with human preference annotations collected from 4 evaluators each; we use the majority vote as the ground-truth label.
It provides neither reference answers nor instance-specific rubrics.
MT-Bench~\cite{10.5555/3666122.3668142} consists of multi-turn dialogues with 3,355 human preference annotations grouped into 1,204 unique pairs.
We take the majority vote and exclude ties, yielding 941 evaluation pairs.
We additionally repurpose \textsc{BiGGen Bench} for pairwise evaluation by comparing all model pairs per prompt and using the higher human score as the preferred response, excluding ties.
This yields 2,792 pairwise evaluation pairs.

\subsection{Training-free Rubrics Generation in Multiple Granularities}
\label{sec:rubric-gen}

We propose a training-free approach to generate evaluation rubrics at two granularity levels without requiring human-annotated data.
We denote the model for generating rubrics as $\mathcal{G}$; in this setting, $\mathcal{G} = \mathcal{J}$ (i.e., the judge model itself generates the rubrics), 
while in Section~\ref{sec:dpo}, we fine-tune a dedicated $\mathcal{G}$ separate from the judge.

\paragraph{Dataset-specific Rubric}
We prompt the judge model to generate a dataset-specific rubric, which is applicable to any instance within a benchmark.
The rubric generation prompt contains no instance-specific information; instead, it asks the model to identify universal dimensions that distinguish good responses from poor ones (e.g., instruction adherence, factual accuracy, completeness, clarity).
This is a cost-efficient setting as the generated rubric is cached and reused for all evaluation instances.

\paragraph{Instance-specific Rubric}
For each instance $x$, we prompt the judge model to generate an instance-specific rubric $\mathcal{R}(x) = \{c_1, \ldots, c_L\}$ 
consisting of $L$ evaluation criteria tailored to the particular instance.
The rubric generation is conditioned only on the instance's input (e.g., the prompt), without seeing any reference responses.
This rubric is then injected into the judge prompt, guiding the model to evaluate responses along the specified dimensions.
The evaluation on response $y$ is conditioned on the instance and generated rubric:
\begin{equation}
  \mathcal{J}(\mathcal{R}(x), x, y).
\end{equation}

Unlike prior rubric generation methods~\cite{li2026rubrichub,ryan2026autometrics}, our approach requires neither reference answers nor human-crafted rubrics as few-shot examples, making it applicable to domains where such resources are unavailable.
Appendix~\ref{sec:appendix-prompts} and~\ref{sec:appendix-examples} provide full prompt templates and examples of generated rubrics.


\subsection{Annotation-free Preference Learning for Rubric Generator}
\label{sec:dpo}

While the training-free approach produces effective rubrics, its quality is bounded by the generator's parametric knowledge.
Empirically, larger models generate better rubrics but at higher cost.
To close this gap, we iteratively fine-tune a small open-source rubric generator via preference learning, still without requiring any human-annotated rubrics or reference answers.
Inspired by Meta-Rewarding~\cite{wu-etal-2025-meta}, we leverage reward signals from a meta-judge to improve rubric generators.
The training pipeline proceeds in five steps: (1) generate diverse rubric candidates, (2) score them using a meta-judge, (3) construct preference pairs, (4) fine-tune the generator with Direct Preference Optimization (DPO)~\cite{NEURIPS2023_a85b405e}, and (5) iterate by using the fine-tuned model as the new base generator.

We use instances from the evaluation benchmarks themselves (Section~\ref{sec:benchmarks}) as training instances.
No human annotations (preference labels or scores) are leaked; the only supervision comes from the meta-judge comparing rubric candidates on intrinsic quality dimensions 
such as correctness, specificity, and coverage.
This is functionally equivalent to best-of-$K$ selection with a meta-judge at inference time, but amortizes the cost by distilling these preferences into the generator offline.
Since the reward signal is rubric-intrinsic and never accesses downstream human judgments, the generator cannot learn to exploit evaluation outcomes.

\noindent
\textbf{Step 1: Rubric Candidate Generation}
For each instance $x$, we generate $K$ diverse rubric candidates $\{\mathcal{R}_1(x), \ldots, \mathcal{R}_K(x)\}$ from the 
base generator $\mathcal{G}$ using high temperature ($\tau{=}0.8$).
This diversity is essential for constructing informative preference pairs.
We use $K{=}8$ candidates per instance.

\noindent
\textbf{Step 2: Meta-Judge Evaluation}
We employ a strong meta-judge model $\mathcal{M}$ (e.g., Claude Sonnet 4) to compare rubric candidates pairwise.
For each pair $(\mathcal{R}_i, \mathcal{R}_j)$, the meta-judge determines which rubric is superior for evaluation, considering:
\begin{compactitem}
  \item \textbf{Specificity}: Are criteria specific?
  \item \textbf{Coverage}: Do criteria address the key aspects an expert would check?
  \item \textbf{Discriminability}: Can criteria distinguish genuinely good responses from superficially good ones?
  \item \textbf{Domain-appropriateness}: Do criteria reflect the required expertise level?
\end{compactitem}
To mitigate positional bias, we evaluate each pair in both orders and count a win only when both orders agree.
From $K{=}8$ candidates ($\binom{8}{2}=28$ possible pairs), we randomly sample 10 pairs per instance to limit computational cost while ensuring each candidate participates in multiple comparisons.

\noindent
\textbf{Step 3: Preference Pair Construction}
We fit a Bradley-Terry model~\cite{19ff28b9-64f9-3656-ba40-08326a05748e} to the meta-judge outcomes, estimating a strength parameter $\gamma_i$ for each candidate by maximizing the likelihood of the observed pairwise comparisons (details in Appendix~\ref{sec:appendix-implementation}).
For each instance, we select the candidate with the highest $\gamma$ as the chosen rubric $\mathcal{R}_w$ and the lowest as the rejected $\mathcal{R}_l$, forming a preference pair $(x,\, \mathcal{R}_w,\, \mathcal{R}_l)$.

\noindent
\textbf{Step 4: DPO Fine-tuning}
We fine-tune the rubric generator $\mathcal{G}_\theta$ using DPO, which directly optimizes the policy to prefer $\mathcal{R}_w$ over $\mathcal{R}_l$:
\begin{equation}
  \mathcal{L}_{\text{DPO}} = -\mathbb{E}\!\left[\log\sigma\!\left(\beta\, r_w - \beta\, r_l\right)\right],
\end{equation}
where $r_i = \log \frac{\pi_\theta(\mathcal{R}_i \mid x)}{\pi_{\text{ref}}(\mathcal{R}_i \mid x)}$ is the log-ratio between the trained policy $\pi_\theta$ and the frozen reference policy $\pi_{\text{ref}}$, and $\beta$ controls the KL constraint strength.


\noindent
\textbf{Step 5: Iterative Training}
The fine-tuned generator from iteration $t$ serves as both the base model and the reference policy for iteration $t{+}1$:
\begin{equation}
  \mathcal{G}^{(t+1)} = \text{DPO}\!\left(\mathcal{G}^{(t)},\; \mathcal{D}^{(t)}\right),
\end{equation}
where $\mathcal{D}^{(t)}$ denotes the preference pairs constructed from rubric candidates generated by $\mathcal{G}^{(t)}$ and scored by the meta-judge.
Steps 1--4 are repeated with the updated generator, allowing it to progressively improve beyond the capability of the base model.
Full implementation details including the hyperparameters are provided in Appendix~\ref{sec:appendix-implementation}.


\section{Experiments}

We evaluate our approach on four benchmarks (Section~\ref{sec:benchmarks}) across four models.
We first compare our training-free rubric generation against existing baselines, 
then evaluate the fine-tuned rubric generator (Section~\ref{sec:exp-dpo}).

\paragraph{Baselines}
\label{sec:baselines}

We compare against Prometheus 2~\cite{kim-etal-2024-prometheus}, an open-source LLM fine-tuned specifically for rubric-based evaluation, 
and three training-free rubric-based methods:
DnA-Eval~\cite{li-etal-2025-dna} decomposes evaluation into 3 weighted aspect-specific questions and aggregates scores per aspect;
CheckEval~\cite{lee-etal-2025-checkeval} generates a fixed checklist of 20 binary questions per benchmark and scores responses by the proportion of affirmative answers;
RubricHub~\cite{li2026rubrichub} generates fine-grained binary criteria grounded in a reference response and aggregates via weighted sum.
All three methods support pointwise evaluation by producing a numeric score per response.
For pairwise benchmarks, we score both responses independently and select the higher-scoring one; 
ties (equal scores) are resolved by random assignment, equivalent to 50\% accuracy on tied pairs.

Additionally, we report results with \textbf{human-crafted rubrics} provided by the benchmark authors.
We evaluate at both dataset-specific~(all benchmarks) and instance-specific granularity (\textsc{BiGGen} only, Section \ref{sec:benchmarks}).
We consider the latter as an oracle upper bound since such fine-grained rubrics are infeasible to obtain at scale in real-world settings.
Appendix~\ref{sec:appendix-baselines} reports further details of the baseline methods.

\begin{table*}[h!]
  \centering
  \small
  \setlength{\tabcolsep}{4pt}
\begin{tabular}{l rrr rr }
\toprule
& \multicolumn{3}{c}{Pairwise}  & \multicolumn{2}{c}{Pointwise} \\
\cmidrule(lr){2-4} \cmidrule(lr){5-6}
& AlpacaEval & MT-Bench &  \textsc{BiGGen} &  HelpSteer2  & \textsc{BiGGen}  \\

\midrule
Baseline \\
~~~Fine-tuned Model \\
~~~~~~Prometheus 8x7B~\cite{kim-etal-2024-prometheus} \\
~~~~~~~~~Dataset-specific                                     & 68.54          & 74.18          & 68.30          & 0.229/0.308          & 0.280/0.307 \\
~~~~~~~~~Instance-specific                  & ---            & ---            & 69.73          & ---                  & 0.366/0.401 \\

\addlinespace

~~~Training-free Approach\\
~~~~~~Llama 3.1 70B\\
~~~~~~~~~CheckEval~\cite{lee-etal-2025-checkeval}  & 58.72          & 71.04          & 64.29          & 0.239/0.324          & 0.318/0.361 \\
~~~~~~~~~DnA-Eval~\cite{li-etal-2025-dna}          & 62.46          & 77.15          & 70.34          & 0.263/0.299          & 0.337/0.353 \\
~~~~~~~~~RubricHub~\cite{li2026rubrichub}          & 62.93          & 80.34          & 69.68          & 0.246/0.292          & 0.318/0.361 \\
~~~~~~~~~Human-crafted (existing) Rubrics \\
~~~~~~~~~~~~Dataset-specific                       & 68.22          & 80.87          & 70.20          & 0.279/0.373          & 0.384/0.382 \\
~~~~~~~~~~~~Instance-specific                      & ---            & ---            & \textbf{76.15} & ---                  & \textbf{0.552}/\textbf{0.594} \\
\addlinespace
~~~~~~Qwen3 14B\\
~~~~~~~~~CheckEval                                 & 60.44          & 75.24          & 65.01          & 0.282/0.392          & 0.343/0.392 \\
~~~~~~~~~DnA-Eval                                  & 69.78          & 78.80          & 71.01          & 0.345/0.403          & 0.431/0.426 \\
~~~~~~~~~RubricHub                                 & 62.93          & 79.76          & 68.28          & 0.286/0.319          & 0.332/0.383 \\
~~~~~~~~~Human-crafted (existing) Rubrics \\
~~~~~~~~~~~~Dataset-specific                       & 70.09          & 79.91          & 72.53          & 0.352/\textbf{0.431} & 0.436/0.450 \\
~~~~~~~~~~~~Instance-specific                      & ---            & ---            & \textbf{76.68} & ---                  & \textbf{0.583}/\textbf{0.609} \\
\addlinespace
~~~~~~Claude Sonnet 4\\
~~~~~~~~~CheckEval                                 & 66.82          & 73.06          & 68.82          & 0.375/0.435          & 0.424/0.455  \\
~~~~~~~~~DnA-Eval                                  & 70.56          & 81.51          & 74.82          & 0.411/0.476          & 0.454/0.489 \\
~~~~~~~~~RubricHub                                 & 64.64          & \textbf{81.72} & 72.22          & 0.239/0.270          & 0.325/0.336 \\
~~~~~~~~~Human-crafted (existing) Rubrics \\
~~~~~~~~~~~~Dataset-specific                       & 70.40          & 81.62          & 74.36          & 0.432/0.483          & 0.476/0.494 \\
~~~~~~~~~~~~Instance-specific                      & ---            & ---            & \textbf{78.33} & ---                  & \textbf{0.584}/\textbf{0.598} \\

\midrule

Rubrics Generation
(Our Approach) \\
~~~Llama 3.1 70B\\
~~~~~~Dataset-specific                             & 67.91               & 79.91                  & 69.73\rlap{$^\ast$}    & 0.284/0.377                      & 0.378/0.361\rlap{$^\ast$} \\
~~~~~~Instance-specific                            &  \textbf{69.78}     & \textbf{80.98}         & 71.74\rlap{$^\ast$}    & \textbf{0.319}/\textbf{0.379}    & 0.420/0.434\rlap{$^\ast$} \\

\addlinespace

~~~Qwen3 14B\\
~~~~~~Dataset-specific                             & 69.47                 & 80.02                  & 72.42\rlap{$^\ast$} & 0.353/0.428                   & 0.440/0.455\rlap{$^\ast$} \\
~~~~~~Instance-specific                            & \textbf{71.96}        & \textbf{80.55}         & 72.96\rlap{$^\ast$} & \textbf{0.374}/0.416          & 0.446/0.462\rlap{$^\ast$} \\

\addlinespace

~~~Claude Sonnet 4\\
~~~~~~Dataset-specific                             &  69.78                & 80.55               & 74.75\rlap{$^\ast$}    & 0.410/0.422\rlap{$^\ast$}         & 0.461/0.453\rlap{$^\ast$} \\
~~~~~~Instance-specific                            &  \textbf{71.65}       & 81.62               & 74.89\rlap{$^\ast$}    & \textbf{0.438}/\textbf{0.488}     & 0.477/0.496\rlap{$^\ast$} \\

\bottomrule
\end{tabular}

  \caption{Evaluation results comparing baselines and our training-free rubric generation. 
  We report human agreement (\%) for pairwise benchmarks and Spearman/Pearson correlation for pointwise benchmarks. 
  The first block includes existing methods (fine-tuned models, training-free baselines, and human-crafted rubrics as oracle upper bounds). 
  Our approach (second block) without training data or human-crafted rubrics yields competitive performance with existing methods (first block).
  \textbf{Bold} marks the best result per judge model.
  Results marked with $^\ast$ are statistically significantly worse than the best existing method in the first block (bootstrap test, $p < 0.05$).
  Our approach obtains comparable results on most benchmarks despite requiring neither model training nor human-annotated data.}
  \label{table:baselines}
\end{table*}
\subsection{Experimental Setting}

\paragraph{Judge Models}
We use four judge models ranging from open-source to proprietary:
Llama 3.1 8B and 70B, Qwen3 14B, and Claude Sonnet 4.
We report main results with the latter three; extra results with Llama 3.1 8B are in Appendix~\ref{sec:appendix-results}.
For pairwise evaluation, we mitigate positional bias~\cite{shi-etal-2025-judging} by evaluating each pair in both orders and counting disagreements as incorrect.

\paragraph{Rubric Generator Model}
\label{sec:training-details}
As discussed in Section~\ref{sec:rubric-gen}, for training-free rubric generation (Section~\ref{sec:exp-training-free}), the judge model itself serves as the rubric generator ($\mathcal{G} = \mathcal{J}$).
For preference learning (Section~\ref{sec:exp-dpo}), we fine-tune Qwen3 14B as a dedicated rubric generator separate from the judge.
Claude Sonnet 4 serves as the meta-judge $\mathcal{M}$ for collecting rubric preferences.
An ablation on different generator models and meta-judges is reported in Appendix~\ref{sec:appendix-ablations}.

\begin{table*}[t]
  \centering
  \small
  \setlength{\tabcolsep}{5pt}
\begin{minipage}[t]{0.47\textwidth}
  \centering
  \begin{tabular}{l rrr}
  \toprule
  & AlpacaEval & MT-Bench & \textsc{BiGGen} \\
  \midrule
  Judge: Llama 3.1 70B \\
  ~~~Highest Training-free$^\dagger$ & 69.78 & 80.98 & 71.74 \\
  ~~~Iteration 1                  & 69.47 & 82.72 & 71.10 \\
  ~~~Iteration 2                  & \textbf{70.72} & \textbf{82.93}\rlap{$^\ast$} & \textbf{71.92} \\
  \addlinespace
  Judge: Qwen3 14B \\
  ~~~Highest Training-free$^\dagger$ & 71.96 & 80.55 & 72.96 \\
  ~~~Iteration 1                  & \textbf{72.40} & 82.62\rlap{$^\ast$} & \textbf{74.67}\rlap{$^\ast$} \\
  ~~~Iteration 2                  & 72.34 & \textbf{82.87}\rlap{$^\ast$} & 74.46 \\
  \addlinespace
  Judge: Claude Sonnet 4 \\
  ~~~Highest Training-free$^\dagger$ & 71.65 & 81.72 & 74.89 \\
  ~~~Iteration 1                  & 72.09 & 83.35 & 76.28 \\
  ~~~Iteration 2                  & \textbf{72.22} & \textbf{83.69}\rlap{$^\ast$} & \textbf{76.96}\rlap{$^\ast$} \\
  \bottomrule
  \end{tabular}
  \subcaption{Pairwise Evaluation (human agreement \%).}
  \label{table:preference_learning_pairwise}
\end{minipage}
\hfill
\begin{minipage}[t]{0.46\textwidth}
  \centering
  \begin{tabular}{l rr}
  \toprule
  & HelpSteer2 & \textsc{BiGGen} \\
  \midrule
  Judge: Llama 3.1 70B \\
  ~~~Highest Training-free$^\dagger$  & 0.319/0.379 & 0.420/0.434 \\
  ~~~Iteration 1                   & 0.326/0.388 & 0.440/0.447\rlap{$^\ast$} \\
  ~~~Iteration 2                   & \textbf{0.344}/\textbf{0.412} & \textbf{0.445}/\textbf{0.452}\rlap{$^\ast$} \\
  \addlinespace
  Judge: Qwen3 14B \\
  ~~~Highest Training-free$^\dagger$ & 0.374/0.416 & \textbf{0.446}/\textbf{0.462} \\
  ~~~Iteration 1                  & 0.394/0.431 & 0.439/0.459 \\
  ~~~Iteration 2                  & \textbf{0.406}/\textbf{0.441}\rlap{$^\ast$} & 0.441/0.461 \\
  \addlinespace
  Judge: Claude Sonnet 4 \\
  ~~~Highest Training-free$^\dagger$ & 0.438/0.488 & 0.477/0.496 \\
  ~~~Iteration 1                  & 0.440/0.471 & 0.490/0.504 \\
  ~~~Iteration 2                  & \textbf{0.464}/\textbf{0.503} & \textbf{0.510}/\textbf{0.521}\rlap{$^\ast$} \\
  \bottomrule
  \end{tabular}
  \subcaption{Pointwise Evaluation (Spearman/Pearson).}
  \label{table:preference_learning_pointwise}
\end{minipage}

  \caption{Results with fine-tuned rubric generator (Qwen3 14B).
  The fine-tuned generator produces \emph{instance-specific} rubrics for each evaluation instance.
  \textbf{Bold} marks the best result per judge model.
  Results with an asterisk ($^\ast$) indicate statistically significant improvement over the highest training-free baseline (bootstrap test, $p < 0.05$).
  $^\dagger$Training-free rows report the best per-model results across all methods in Table~\ref{table:baselines}, excluding oracle human-crafted instance-specific rubrics.
  The fine-tuned generator consistently outperforms all existing methods and narrows the gap to oracle human-crafted rubrics.}
  \label{table:preference_learning}
\end{table*}


\subsection{Results with Baselines and Training-Free Rubric Generation}
\label{sec:exp-training-free}

Table~\ref{table:baselines} presents results comparing our training-free rubric generation against baselines and human-crafted rubrics, reporting human agreement (\%) for pairwise benchmarks and Spearman/Pearson correlation for pointwise benchmarks.

\paragraph{Existing Baselines}
Among the three existing methods, DnA-Eval consistently achieves the strongest results, 
outperforming CheckEval and RubricHub on most benchmarks---the only exception is MT-Bench.
Prometheus 2 (8$\times$7B) underperforms the best training-free methods despite being explicitly trained for rubric-based evaluation, 
suggesting that general LLMs with high-quality rubrics can surpass specialized LLM judges.

Despite being the strongest baseline,
DnA-Eval still falls short of human-crafted dataset-specific rubrics on most benchmarks (e.g., 69.78\% vs.\ 70.09\% on AlpacaEval, 78.80\% vs.\ 79.91\% on MT-Bench, both with Qwen3 14B).
We attribute this gap to \emph{annotation alignment}: human-crafted rubrics mirror the exact evaluation protocol used by annotators 
(e.g., HelpSteer2's rubric uses helpfulness, correctness, and coherence, 
the same dimensions averaged to compute the ground-truth score), 
whereas existing methods generate ad-hoc aspects that may not align with the annotation criteria.

Human-crafted instance-specific rubrics (available only for \textsc{BiGGen}) further outperform all automated methods by a large margin (78.33\% vs.\ 74.82\%, 0.584 vs.\ 0.454 with Claude Sonnet 4), 
highlighting the value of fine-grained, expert-crafted rubrics.
However, such rubrics require domain expertise that is expensive and often infeasible to collect at scale.

\paragraph{Our Approach}
Our training-free rubric generation achieves competitive and even superior results to the existing methods without any training data, human-crafted rubrics, or reference answers.
Generated instance-specific rubrics consistently outperform dataset-specific across all judge models, 
again confirming that fine-grained rubrics work better even without reference answers.
Notably, our approach achieves state-of-the-art results on AlpacaEval and HelpSteer2 with all models.

The majority of our results are \textbf{not} statistically significantly worse than the best existing method (reported in Table~\ref{table:baselines} without an asterisk ($^\ast$); bootstrap test, $p > 0.05$).
The only significant gaps appear on \textsc{BiGGen} with human-crafted instance-specific rubrics, which is considered an oracle upper bound.
Appendix~\ref{sec:appendix-significance} reports more details including the significance testing. 

\subsection{Results with Fine-tuned Rubric Generator}
\label{sec:exp-dpo}

We apply the preference fine-tuned rubric generator exclusively for \emph{instance-specific} rubric generation.
Dataset-specific rubrics, in principle, would benefit less from preference learning since their quality is already high with the base models.
We train separate generators for pairwise and pointwise evaluation, 
as the two settings demand different rubric properties: 
pairwise rubrics must highlight dimensions that differentiate between two responses, 
while pointwise rubrics must define criteria for absolute scoring.
Table~\ref{table:preference_learning} presents results using rubrics from the fine-tuned Qwen3 14B generator over two iterations.

\begin{table}[t!]
  \centering
  \small
  \begin{tabular}{l rrr}
\toprule
& Base & Iter 1 & Iter 2 \\
\midrule

\multicolumn{4}{l}{\textit{Avg.\ $\Delta$ vs.\ Base}} \\
~~Pairwise agreement & --- & +1.15 & +1.53 \\
~~Pointwise Spearman & --- & +0.017 & +0.030 \\
\addlinespace
\multicolumn{4}{l}{\textit{Structural properties (pairwise)}} \\
~~Avg.\ length (chars) & 621 & 630 & 634 \\
~~Avg.\ criteria / rubric & 4.86 & 4.80 & 4.89 \\
~~Prescriptive (\%) $\downarrow$ & 47.2 & 39.5 & 24.9 \\
\addlinespace
\multicolumn{4}{l}{\textit{Structural properties (pointwise)}} \\
~~Avg.\ length (chars) & 637 & 619 & 699 \\
~~Avg.\ criteria / rubric & 4.9 & 4.9 & 5.0 \\
~~Prescriptive (\%) $\downarrow$ & 38.6 & 33.5 & 0.9 \\
\addlinespace
\multicolumn{4}{l}{\textit{Meta-judge statistics (pairwise)}} \\
~~Best-candidate win rate $\uparrow$ & 0.30 & 0.53 & --- \\
~~Preference margin $\downarrow$ & 0.80 & 0.67 & --- \\
\addlinespace
\multicolumn{4}{l}{\textit{Same-instance divergence (pairwise)}} \\
~~Word Jaccard vs.\ Base $\downarrow$ & --- & 0.536 & 0.498 \\

\bottomrule
\end{tabular}

  \caption{Rubric properties across iterations (Qwen3 14B generator).
  Prescriptive (\%) counts criteria beginning with ``The response...''.
  Meta-judge statistics are computed from the $K{=}8$ candidates sampled during reward signal collection: best-candidate win rate is the fraction of comparisons won by the top candidate; preference margin is the win-rate gap between chosen and rejected candidates in preference pairs.
  Same-instance divergence measures word overlap of the same instance's rubric across iterations.}
  \label{table:rubric_analysis}
\end{table}

The fine-tuned generator consistently improves over training-free rubric generation across all benchmarks.
Many of these improvements are statistically significant (with an asterisk $^\ast$, e.g., 83.69\% vs.\ 81.72\% on MT-Bench and 76.96\% vs.\ 74.89\% on \textsc{BiGGen} with Claude Sonnet 4).
Importantly, the rubrics from the fine-tuned generator outperform most existing methods on every benchmark.
Again, the only exception is the remaining gap to the oracle on \textsc{BiGGen}, though it narrows with stronger judges (4.23 to 1.37 percentage points from Llama 3.1 70B to Claude Sonnet 4).

Notably, the fine-tuned Qwen3 14B generator outperforms Claude Sonnet 4 at rubric generation despite being a much smaller model: rubrics from Qwen3 14B achieve 83.69\% on MT-Bench and 76.96\% on \textsc{BiGGen} (with Claude as judge), compared to 81.62\% and 74.89\% when Claude generates its own rubrics (Table~\ref{table:baselines}).
Moreover, the fine-tuned Qwen3 14B paired with Qwen3 14B as judge (72.34\% AlpacaEval, 82.87\% MT-Bench) outperforms Claude Sonnet 4 as both generator and judge (71.65\%, 81.62\%),
demonstrating that overall evaluation quality can be decoupled from judge capability.
This suggests that under limited computational budgets, investing in a specialized rubric generator can yield greater returns than scaling up the judge model alone.

Furthermore, the iterative training procedure (Section~\ref{sec:dpo}) yields gains across both iterations.
Compared to the base Qwen3 14B generator, the first iteration improves pairwise agreement by +1.15 percentage points on average, and the second iteration further increases this to +1.53 (Table~\ref{table:rubric_analysis}).
We observe a similar trend on pointwise evaluation (Spearman improved by +0.017 after the first iteration, +0.030 after the second).
However, the marginal gain diminishes from Iteration~1 to Iteration~2 (+1.15 vs.\ +0.38 for pairwise; +0.017 vs.\ +0.013 for Spearman),
suggesting potential performance convergence. We do not explore further iterations due to computational constraints.

Finally, we conduct several ablation studies, including fine-tuning a rubric generator jointly on both pairwise and pointwise instances, 
using a different model as the meta-judge for preference collection, and fine-tuning a smaller rubric generator model (Llama 3.1 8B).
Due to space limitations, these ablations are reported in Appendix~\ref{sec:appendix-ablations}.

\paragraph{Qualitative Analysis}

We compare rubrics produced by Qwen3 14B as the base, Iteration~1, and Iteration~2 models (Table~\ref{table:rubric_analysis}).
Rubric length and criteria count remain stable across iterations, indicating that improvement is qualitative rather than structural.
The key shift is in rubric \emph{framing}: the base model uses prescriptive criteria (``The response clearly explains \dots'') while 
Iteration~2 reframes around evaluation dimensions (``Accuracy of \dots'', ``Clarity and coherence in \dots'').
Prescriptive criteria drop from 47.2\% to 24.9\% for pairwise rubrics, with an even sharper decline for pointwise (38.6\%$\to$0.9\%).
The same-instance divergence between base and Iteration~2 is only 0.498 (i.e., fewer than half of the words are shared), confirming that preference learning substantially rewrites rubrics rather than merely paraphrasing them.
This reframing provides stronger discriminative signal for pairwise comparison, explaining the larger gains on pairwise benchmarks.
The meta-judge statistics further support this: the best-candidate win rate increases from 0.30 (base) to 0.53 (Iteration~1), indicating that stronger rubric candidates emerge as training progresses.
We provide extra analysis and three case studies to further explain the observed performance gains in Appendix~\ref{sec:appendix-qualitative}.



\section{Conclusion}

We presented an approach to improve LLM-as-a-Judge through automatically generated evaluation rubrics, requiring no human-annotated data or reference answers.
Our training-free rubric generation achieves competitive performance with existing methods across four benchmarks.
We further proposed a preference learning method to iteratively fine-tune a rubric generator via meta-judge feedback,
consistently outperforming all existing baselines.
Notably, a fine-tuned Qwen3 14B generator produces higher-quality rubrics than Claude Sonnet 4, 
demonstrating that evaluation quality can be decoupled from judge model scale through specialized rubric generation.

Future work includes incorporating reference answers when available to improve rubric specificity, evaluating cross-domain generalization of the fine-tuned generator, and exploring alternative reward signals beyond meta-judge preferences.

\section*{Limitations}

Our approach requires multiple LLM calls per evaluation instance (rubric generation followed by judgment), increasing computational cost compared to direct scoring.
The preference learning procedure relies on a capable meta-judge for reward signal collection, which may limit accessibility in resource-constrained settings.
We explored only two iterations due to computational constraints; the sustained improvement trend suggests further iterations may yield additional gains, but potential saturation or degradation remains unexplored.
We do not investigate the trade-off between rubric generator models and LLM judge models, which would be beneficial for practical usage with limited resources.
We train and evaluate on the same benchmarks; an out-of-domain evaluation (training on one benchmark and testing on another) would better test generalization, but few meta-evaluation benchmarks with reliable human judgments exist beyond the ones we use.
We do not compare the fine-tuned generator against best-of-$K$ selection at inference time (generating $K$ rubrics and selecting the best via meta-judge), which would isolate the benefit of amortizing preferences into the generator versus paying the cost at inference.
We only evaluate under pairwise and pointwise settings, omitting reference-based evaluation where the judge compares a response against a gold answer. This is partly because reference answers are scarce in many real-world datasets, but extending our rubric generation approach to the reference-based setting remains future work.
Finally, our evaluation is limited to mostly general domain benchmarks. An extension to domain-specific benchmarks (e.g., medical, legal, finance) would be valuable.

\section*{Ethics Statement}

We used AI-based assistance to improve the language and clarity of this paper, and as a coding aid during implementation. All research ideas, experimental design, analysis, and conclusions are entirely the work of the authors.



\bibliography{custom}

\appendix

\section{Prompt Templates}
\label{sec:appendix-prompts}

We list all prompt templates used in our system. Each box shows the system and user messages sent to the model.

\noindent\textbf{Rubric Generation Prompts}

\begin{tcolorbox}[
  width=\linewidth, colback=white, colframe=gray!50, arc=1.5mm, boxrule=0.5pt,
  title={\textbf{\textsf{Instance-specific rubric generation (pointwise)}}},
  fonttitle=\small, fontupper=\small\ttfamily, fontlower=\small\ttfamily, bottom=4pt, top=4pt, left=4pt, right=4pt
]
{\normalfont\bfseries System:} You are an expert evaluator designing assessment criteria. Given a task prompt, create a concise evaluation rubric with 3--5 specific criteria for scoring the quality of responses to this prompt on a 1--5 scale.

Focus on what makes a response good or bad FOR THIS SPECIFIC TASK. Consider the task type (factual Q\&A, creative writing, coding, reasoning, summarization, etc.) and what a high-quality answer requires.

Output ONLY a numbered list of criteria. Each criterion should be one clear sentence.

\tcblower
{\normalfont\bfseries User:} [Prompt]\\
\{prompt\}\\[0.5em]
Write the evaluation rubric for scoring responses to the above prompt.
\end{tcolorbox}

\begin{tcolorbox}[
  width=\linewidth, colback=white, colframe=gray!50, arc=1.5mm, boxrule=0.5pt,
  title={\textbf{\textsf{Dataset-specific rubric generation (pointwise)}}},
  fonttitle=\small, fontupper=\small\ttfamily, fontlower=\small\ttfamily, bottom=4pt, top=4pt, left=4pt, right=4pt
]
{\normalfont\bfseries System:} You are an expert evaluator designing assessment criteria for scoring AI assistant responses. Your task is to create a general-purpose evaluation rubric that can be applied to ANY type of prompt-response pair, where each response is scored individually on a 1--5 scale.

The rubric should cover the key dimensions that distinguish a good AI response from a poor one, regardless of the specific task.

Output ONLY a numbered list of 3--5 criteria. Each criterion should be one clear sentence describing what to evaluate.

\tcblower
{\normalfont\bfseries User:} Create a general-purpose evaluation rubric for scoring individual AI assistant responses on a 1--5 scale. The rubric should be applicable to any type of task (factual Q\&A, creative writing, coding, reasoning, summarization, advice, etc.) without any task-specific or instance-specific information.

Write the evaluation rubric.
\end{tcolorbox}

\begin{tcolorbox}[
  width=\linewidth, colback=white, colframe=gray!50, arc=1.5mm, boxrule=0.5pt,
  title={\textbf{\textsf{Instance-specific rubric generation (pairwise)}}},
  fonttitle=\small, fontupper=\small\ttfamily, fontlower=\small\ttfamily, bottom=4pt, top=4pt, left=4pt, right=4pt
]
{\normalfont\bfseries System:} You are an expert evaluator designing assessment criteria. Given a task instruction, create a concise evaluation rubric with 3--5 specific criteria for judging the quality of responses to this instruction.

Focus on what makes a response good or bad FOR THIS SPECIFIC TASK. Consider the task type (factual Q\&A, creative writing, coding, reasoning, summarization, etc.) and what a high-quality answer requires.

Output ONLY a numbered list of criteria. Each criterion should be one clear sentence.

\tcblower
{\normalfont\bfseries User:} [Instruction]\\
\{instruction\}\\[0.5em]
Write the evaluation rubric for judging responses to the above instruction.
\end{tcolorbox}

\begin{tcolorbox}[
  width=\linewidth, colback=white, colframe=gray!50, arc=1.5mm, boxrule=0.5pt,
  title={\textbf{\textsf{Dataset-specific rubric generation (pairwise)}}},
  fonttitle=\small, fontupper=\small\ttfamily, fontlower=\small\ttfamily, bottom=4pt, top=4pt, left=4pt, right=4pt
]
{\normalfont\bfseries System:} You are an expert evaluator designing assessment criteria for comparing AI assistant responses. Your task is to create a general-purpose evaluation rubric that can be applied to ANY type of instruction-response pair.

The rubric should cover the key dimensions that distinguish a good AI response from a poor one, regardless of the specific task.

Output ONLY a numbered list of 3--5 criteria. Each criterion should be one clear sentence describing what to evaluate.

\tcblower
{\normalfont\bfseries User:} Create a general-purpose evaluation rubric for judging pairwise comparisons of AI assistant responses. The rubric should be applicable to any type of task (factual Q\&A, creative writing, coding, reasoning, summarization, advice, etc.) without any task-specific or instance-specific information.

Write the evaluation rubric.
\end{tcolorbox}

\noindent\textbf{Judge Prompts}

\begin{tcolorbox}[
  width=\linewidth, colback=white, colframe=gray!50, arc=1.5mm, boxrule=0.5pt,
  title={\textbf{\textsf{Pointwise judge}}},
  fonttitle=\small, fontupper=\small\ttfamily, fontlower=\small\ttfamily, bottom=4pt, top=4pt, left=4pt, right=4pt
]
{\normalfont\bfseries System:} You are a helpful assistant that evaluates the quality of AI responses. You will be given a prompt, an evaluation rubric, and an AI response. Use the rubric criteria to assess the response and assign a score from 1 to 5.

Scoring scale:\\
~~1 = Poor -- The response fails to address the prompt or has critical issues\\
~~2 = Below Average -- The response partially addresses the prompt but has significant shortcomings\\
~~3 = Average -- The response adequately addresses the prompt with some room for improvement\\
~~4 = Good -- The response effectively addresses the prompt with minor issues\\
~~5 = Excellent -- The response thoroughly and skillfully addresses the prompt

You must output ONLY a single JSON object with two keys:\\
- "score": an integer from 1 to 5\\
- "reason": a brief one-sentence explanation referencing the rubric criteria

\tcblower
{\normalfont\bfseries User:} [Prompt]\\
\{prompt\}\\[0.5em]
[Evaluation Rubric]\\
\{rubric\}\\[0.5em]
[Response]\\
\{response\}\\[0.5em]
Based on the rubric, score this response from 1 to 5. Output only the JSON object.
\end{tcolorbox}

\begin{tcolorbox}[
  width=\linewidth, colback=white, colframe=gray!50, arc=1.5mm, boxrule=0.5pt,
  title={\textbf{\textsf{Pairwise judge}}},
  fonttitle=\small, fontupper=\small\ttfamily, fontlower=\small\ttfamily, bottom=4pt, top=4pt, left=4pt, right=4pt
]
{\normalfont\bfseries System:} You are a helpful assistant that evaluates the quality of AI responses. You will be given an instruction, an evaluation rubric tailored to that instruction, and two responses (Response A and Response B). Use the rubric criteria to compare the responses and decide which one is better.

You must output ONLY a single JSON object with two keys:\\
- "winner": either "A" or "B" (the letter of the better response)\\
- "reason": a brief one-sentence explanation referencing the rubric criteria

\tcblower
{\normalfont\bfseries User:} [Instruction]\\
\{instruction\}\\[0.5em]
[Evaluation Rubric]\\
\{rubric\}\\[0.5em]
[Response A]\\
\{response\_a\}\\[0.5em]
[Response B]\\
\{response\_b\}\\[0.5em]
Based on the rubric, which response is better? Output only the JSON object.
\end{tcolorbox}

\noindent\textbf{Meta-Judge and Reward Signal Collection Prompts}

\begin{tcolorbox}[
  width=\linewidth, colback=white, colframe=gray!50, arc=1.5mm, boxrule=0.5pt,
  title={\textbf{\textsf{Meta-judge (rubric comparison)}}},
  fonttitle=\small, fontupper=\small\ttfamily, fontlower=\small\ttfamily, bottom=4pt, top=4pt, left=4pt, right=4pt
]
{\normalfont\bfseries System:} You are a meta-evaluator that assesses the quality of evaluation rubrics. You will be given a task prompt and two candidate rubrics (Rubric A and Rubric B). Determine which rubric would be more effective for evaluating AI responses to the given prompt.

Consider these dimensions:\\
1. Specificity: Are criteria concrete and testable (yes/no answerable), or vague and subjective?\\
2. Coverage: Do criteria address the key aspects an expert would check?\\
3. Discriminability: Can these criteria distinguish genuinely good responses from superficially good ones?\\
4. Domain-appropriateness: Do criteria reflect the expertise level the task requires?

Output ONLY a JSON object:\\
\{"winner": "A" or "B", "reason": "brief explanation"\}

\tcblower
{\normalfont\bfseries User:} [Task Prompt]\\
\{prompt\}\\[0.5em]
[Rubric A]\\
\{rubric\_a\}\\[0.5em]
[Rubric B]\\
\{rubric\_b\}\\[0.5em]
Which rubric is better for evaluating responses to this task? Output only the JSON object.
\end{tcolorbox}

\begin{tcolorbox}[
  width=\linewidth, colback=white, colframe=gray!50, arc=1.5mm, boxrule=0.5pt,
  title={\textbf{\textsf{Rubric candidate generation (for reward collection)}}},
  fonttitle=\small, fontupper=\small\ttfamily, fontlower=\small\ttfamily, bottom=4pt, top=4pt, left=4pt, right=4pt
]
{\normalfont\bfseries System:} You are an expert evaluator designing assessment criteria for AI responses. Given a task prompt, generate specific evaluation criteria that can each be answered with a clear YES or NO when checking a response.

Guidelines:\\
- Each criterion must describe a concrete, observable requirement -- not an abstract quality dimension like "clarity" or "accuracy."\\
- Criteria should be specific to THIS prompt, capturing what a domain expert would check for in a high-quality answer.\\
- Include both content requirements (what must be covered) and safety/quality guards (what must be avoided).\\
- Assign each criterion a weight reflecting its importance:\\
~~Critical -- essential; failure indicates a serious flaw\\
~~Major~~~~-- important for a good response but not essential\\
~~Minor~~~~-- desirable but lower impact

Output ONLY a JSON array of 5--10 objects with keys "criterion" and "weight".

\tcblower
{\normalfont\bfseries User:} [Prompt]\\
\{prompt\}\\[0.5em]
Generate 5--10 specific, binary-testable evaluation criteria for responses to the above prompt. Output only the JSON array.
\end{tcolorbox}

\section{Example Rubrics}
\label{sec:appendix-examples}

We provide examples of generated rubrics at both granularity levels. Figure~\ref{fig:dataset_rubric} shows a dataset-specific rubric applicable to any instance, and Figure~\ref{fig:instance_rubric} shows an instance-specific rubric tailored to a particular task.

\begin{figure}[t]
  \centering
  \begin{tcolorbox}[
    width=\linewidth,
    colback=white,
    colframe=gray!50,
    colbacktitle=blue!20,
    coltitle=black,
    arc=1.5mm,
    boxrule=0.5pt,
    title={\textbf{\textsf{Dataset-specific rubric (generated)}}},
    fonttitle=\small,
    fontupper=\small\textsf{}
  ]
  \begin{compactitem}
\item Accuracy and Correctness: Evaluate whether the response provides factually correct information, follows logical reasoning, and directly addresses what was asked without significant errors or misconceptions.

\item Completeness and Relevance: Assess how thoroughly the response covers the key aspects of the instruction while staying focused on the topic and avoiding unnecessary tangents or omissions.

\item Clarity and Organization: Judge how well-structured, coherent, and easy to understand the response is, including appropriate use of formatting, logical flow, and clear communication.

\item Helpfulness and Practical Value: Determine how useful the response would be to the human, considering whether it provides actionable information, appropriate depth, and genuinely assists with the stated need.

\item Safety and Appropriateness: Evaluate whether the response avoids harmful, biased, or inappropriate content while maintaining ethical standards and respecting relevant guidelines.
\end{compactitem}

\end{tcolorbox}
  \caption{Example dataset-specific evaluation rubric generated by Claude Sonnet 4.}
  \label{fig:dataset_rubric}
\end{figure}

\begin{figure}[t]
  \centering
  \begin{tcolorbox}[
    width=\linewidth,
    colback=white,
    colframe=gray!50,
    colbacktitle=blue!20,
    coltitle=black,
    titlerule=0.5pt,
    arc=1.5mm,
    boxrule=0.5pt,
    title={\textbf{\textsf{Instance-specific rubric (generated)}}},
    fonttitle=\small,
    fontupper=\small\textsf{},
    fontlower=\small\textsf{},
    bottom=4pt, top=4pt,
    left=4pt, right=4pt
  ]

  \textbf{Instruction:} \\[0.5em]
  The sentence you are given might be too wordy, complicated, or unclear. Rewrite the sentence and make your writing clearer by keeping it concise. Whenever possible, break complex sentences into multiple sentences and eliminate unnecessary words.

  \textbf{Instance:} \\[0.5em]
  If you have any questions about my rate or if you find it necessary to increase or decrease the scope for this project, please let me know.

   \tcblower

  \textbf{Rubric:}
  \begin{compactitem}
      \item The rewritten sentence(s) should be significantly more concise than the original, eliminating unnecessary or redundant words.
      \item Complex ideas should be broken into multiple clear, simple sentences rather than one complicated sentence.
      \item The core meaning and intent of the original message must be preserved without any loss of important information.
      \item The language should be more direct and straightforward, avoiding wordy phrases and unnecessary qualifiers.
      \item The final result should improve readability and clarity compared to the original sentence.
  \end{compactitem}
  \end{tcolorbox}
  \caption{Example instance-specific evaluation rubric generated by Claude Sonnet 4.}
  \label{fig:instance_rubric}
\end{figure}

\section{Implementation Details}
\label{sec:appendix-implementation}

\paragraph{Model Inference}
Llama 3.1 8B Instruct, Qwen3 14B, and Llama 3.1 70B Instruct are served on a locally hosted NVIDIA H100 GPU (94GB) via vLLM~\cite{10.1145/3600006.3613165}.
Claude Sonnet 4 is accessed via the AWS Bedrock API.
All judge and rubric generation calls use greedy decoding (temperature 0) with a maximum of 256 output tokens for judging and 1024 tokens for rubric generation.
During reward signal collection, rubric candidates are sampled with temperature 0.8 to encourage diversity.

\paragraph{Preference Fine-tuning}
We fine-tune both Llama 3.1 8B Instruct and Qwen3 14B using QLoRA~\cite{dettmers2023qlora} with 4-bit quantization via DPO.
The same hyperparameters are used for both models (Table~\ref{table:hyperparams}).

\begin{table}[h]
  \centering
  \small
  \begin{tabular}{l l}
    \toprule
    Hyperparameter & Value \\
    \midrule
    Base models & Llama 3.1 8B / Qwen3 14B \\
    LoRA rank & 64 \\
    LoRA $\alpha$ & 128 \\
    LoRA dropout & 0.05 \\
    Quantization & 4-bit \\
    DPO $\beta$ & 0.1 \\
    Learning rate & $5 \times 10^{-6}$ \\
    Batch size (effective) & 16 \\
    Epochs & 3 \\
    Max sequence length & 2048 \\
    Optimizer & AdamW \\
    \bottomrule
  \end{tabular}
  \caption{Preference fine-tuning hyperparameters for the rubric generator.}
  \label{table:hyperparams}
\end{table}

\paragraph{Meta-Judge Configuration}
We use Claude Sonnet 4 as the meta-judge model.
For each instance, we generate $K{=}8$ rubric candidates with temperature $\tau{=}0.8$ and sample up to 10 pairwise comparisons.
Positional bias is mitigated by evaluating each pair in both presentation orders.

\paragraph{Bradley-Terry Ranking}
We fit a Bradley-Terry model to the meta-judge outcomes. The model assumes each candidate $\mathcal{R}_i$ has a latent strength $\gamma_i > 0$, with the probability of $\mathcal{R}_i$ being preferred over $\mathcal{R}_j$ given by:
\begin{equation}
  P(\mathcal{R}_i \succ \mathcal{R}_j) = \frac{\gamma_i}{\gamma_i + \gamma_j}.
\end{equation}
We estimate $\gamma_1, \ldots, \gamma_K$ by maximizing the log-likelihood of the observed outcomes:
\begin{equation}
  \mathcal{L} = \sum_{(i,j) \in \mathcal{O}} \log \frac{\gamma_i}{\gamma_i + \gamma_j},
\end{equation}
where $\mathcal{O}$ is the set of observed preferences (each entry $(i,j)$ indicates $\mathcal{R}_i$ was preferred over $\mathcal{R}_j$ by the meta-judge).
The candidate with the highest $\gamma$ becomes the chosen rubric and the lowest becomes the rejected, forming a preference pair for DPO training.

\paragraph{Compute}
All fine-tuning runs use a single NVIDIA H100 GPU (94GB). Each run completes in approximately 1--2 hours for 3 epochs.

\section{Baseline Method Details}
\label{sec:appendix-baselines}

\paragraph{Prometheus 2}
Prometheus 2~\cite{kim-etal-2024-prometheus} is a family of open-source LLMs (7B and 8$\times$7B MoE) fine-tuned specifically for evaluation tasks.
The models are trained on a large corpus of evaluation data encompassing both pairwise and pointwise formats, with rubric-conditioned scoring.
Given an instance and a rubric, Prometheus 2 directly produces a score (pointwise) or preference judgment (pairwise) without requiring multi-step decomposition.
We report results with the 8$\times$7B variant using both dataset-specific and instance-specific rubrics (where available).

\paragraph{DnA-Eval}
Decomposition and Aggregation~\cite{li-etal-2025-dna} decomposes evaluation into multiple aspect-specific questions with importance weights.
For each instance $x$, the method generates $K$ evaluation aspects $\{a_1, \ldots, a_K\}$ and corresponding importance weights $\{w_1, \ldots, w_K\}$ (with $\sum_k w_k = 1$) using the judge model itself.
For pairwise evaluation, each response is scored on each aspect independently on a 1--10 scale, yielding aspect scores $\{s_k^{(1)}\}$ and $\{s_k^{(2)}\}$, and the final preference is determined by the weighted aggregation:
\begin{equation}
  \hat{y} = \arg\max_{i \in \{1,2\}} \sum_{k=1}^{K} w_k \cdot s_k^{(i)}.
\end{equation}
For pointwise evaluation, the judge score is $\sum_k w_k \cdot s_k$.
We use $K{=}3$ aspects following the original paper.

\paragraph{CheckEval}
CheckEval~\cite{lee-etal-2025-checkeval} decomposes evaluation into a fixed checklist of binary (Yes/No) questions.
A set of $N$ questions $\{q_1, \ldots, q_N\}$ is generated once per benchmark (not per instance) and applied uniformly to all items.
For each response, the judge answers all $N$ questions, and the score is computed as the proportion of affirmative answers:
\begin{equation}
  s = \frac{1}{N}\sum_{n=1}^{N} \mathbbm{1}[\text{answer}(q_n) = \text{Yes}].
\end{equation}
For pairwise comparison, both responses are scored independently and the higher-scoring response is preferred.
We use $N{=}20$ questions.

\paragraph{RubricHub}
RubricHub~\cite{li2026rubrichub} generates fine-grained, weighted binary criteria grounded in a reference response.
The pipeline consists of four stages: (1) generate a reference response $r$ for instance $x$; (2) generate a set of rubric criteria $\{(t_i, d_i, w_i)\}_{i=1}^{M}$ (each with title $t_i$, description $d_i$, and importance weight $w_i \in [1, 10]$) conditioned on both $x$ and $r$; (3) grade the candidate response on each criterion independently, producing binary verdicts $b_i \in \{0, 1\}$; and (4) aggregate via weighted sum:
\begin{equation}
  s = \frac{\sum_{i=1}^{M} w_i \cdot b_i}{\sum_{i=1}^{M} w_i}.
\end{equation}
References and rubrics are cached per instance so that repeated instances share the same evaluation criteria.

\section{Additional Results with Llama 3.1 8B as Judge}
\label{sec:appendix-results}

Table~\ref{table:baselines_appendix} reports baselines and our training-free rubric generation results with Llama 3.1 8B as the judge model.

\begin{table*}[t]
  \centering
  \small
  \begin{tabular}{l rrr rr }
\toprule
& \multicolumn{3}{c}{Pairwise}  & \multicolumn{2}{c}{Pointwise} \\
\cmidrule(lr){2-4} \cmidrule(lr){5-6}
& AlpacaEval & MT-Bench &  \textsc{BiGGen} &  HelpSteer2  & \textsc{BiGGen}  \\

\midrule
Prometheus~\cite{kim-etal-2024-prometheus} \\
~~~7B                                              & 62.62          & 72.16          & 67.59          & 0.143/0.215          & 0.254/0.281         \\
~~~7B (instance-specific)                          & ---            & ---            & 66.19          & ---                  & 0.328/0.356         \\

\midrule
Baseline (Judge: Llama 3.1 8B) \\
~~~CheckEval~\cite{lee-etal-2025-checkeval}        & 56.54          & 70.56          & 63.29          & \textbf{0.195}/\textbf{0.256} & 0.250/0.280 \\
~~~DnA-Eval~\cite{li-etal-2025-dna}                & 59.81          & 69.45          & 65.69          & 0.168/0.207          & 0.303/0.240 \\
~~~RubricHub~\cite{li2026rubrichub}                & 58.26          & 73.86          & 66.94          & 0.180/0.177          & 0.239/0.252 \\
~~~Human-crafted Rubrics \\
~~~~~~Dataset-specific                             & 65.11          &  73.22         & 64.53          & 0.152/0.226          & 0.292/0.280 \\
~~~~~~Instance-specific                            & ---            & ---            & \textbf{70.31} & ---                  & \textbf{0.420}/\textbf{0.483} \\

\midrule

Our Approach (Judge: Llama 3.1 8B) \\
~~~Dataset-specific                                & 64.80          & 73.01          & 64.61          & 0.135/0.224          & 0.276/0.277 \\
~~~Instance-specific                               & 65.98          & 74.39          & 67.30          & \textbf{0.224}/\textbf{0.284} & 0.297/0.317 \\

\bottomrule
\end{tabular}

  \caption{Baselines and training-free rubric generation results with Llama 3.1 8B as the judge. The trends are consistent with the main results in Table~\ref{table:baselines}: our instance-specific rubrics outperform dataset-specific ones, and the overall performance is lower due to the weaker judge model.}
  \label{table:baselines_appendix}
\end{table*}

\section{Statistical Significance Testing}
\label{sec:appendix-significance}

We test statistical significance between our approach and the best existing method for each benchmark--judge combination using the following procedures.

\paragraph{Pairwise Benchmarks}
For pairwise benchmarks, each method produces a binary per-instance outcome (correct or incorrect agreement with the human majority vote).
We apply a paired bootstrap test with 10,000 resamples: in each iteration, we resample instances with replacement and compute the accuracy difference between the two methods.
The two-sided $p$-value is computed as twice the proportion of bootstrap samples where the observed difference reverses sign.
We additionally report McNemar's test with continuity correction as a complementary check on paired binary outcomes.

\paragraph{Pointwise Benchmarks}
For pointwise benchmarks, we compare Spearman correlations via a paired bootstrap procedure.
In each of 10,000 iterations, we resample (instance, score) pairs with replacement and recompute both methods' correlations with human scores.
The $p$-value is derived from the proportion of bootstrap samples where the correlation difference reverses sign (two-sided).

Results marked with $^\ast$ in Table~\ref{table:baselines} indicate cases where our method is statistically significantly worse than the best existing method at $p < 0.05$.

\section{Ablation Studies}
\label{sec:appendix-ablations}

We ablate three design choices. The first two use the Llama 3.1 8B rubric generator (Iteration~1) evaluated with Claude Sonnet 4 as the judge (Table~\ref{table:ablations}); the third varies the generator model size.

\begin{table*}[h]
  \centering
  \small
  \begin{tabular}{l rrr rr}
    \toprule
    & \multicolumn{3}{c}{Pairwise} & \multicolumn{2}{c}{Pointwise} \\
    \cmidrule(lr){2-4} \cmidrule(lr){5-6}
    & AlpacaEval & MT-Bench & \textsc{BiGGen} & HelpSteer2 & \textsc{BiGGen} \\
    \midrule
    Separate (main) & \textbf{70.09} & 82.14 & 74.25 & 0.392/0.428 & 0.434/0.441 \\
    Joint training & 68.85 & 81.41 & 74.21 & 0.389/0.420 & 0.441/0.448 \\
    \addlinespace
    Meta-judge: Claude Sonnet 4 & \textbf{70.09} & 82.14 & 74.25 & --- & --- \\
    Meta-judge: Llama 3.1 70B & 67.50 & \textbf{82.55} & \textbf{74.46} & --- & --- \\
    \bottomrule
  \end{tabular}

  \caption{Ablation studies with Llama 3.1 8B rubric generator (Iteration~1), evaluated with Claude Sonnet 4 as judge. ``Separate'' trains distinct pairwise and pointwise generators; ``Joint'' trains one model on both. Meta-judge rows compare Claude Sonnet 4 vs.\ Llama 3.1 70B for reward signal collection.}
  \label{table:ablations}
\end{table*}

\paragraph{Joint Pairwise--Pointwise Training}
In our main experiments, we train separate rubric generators for pairwise and pointwise evaluation.
Here we ablate this choice by training a single model on the combined preference data.
Joint training degrades pairwise performance (70.09$\to$68.85 on AlpacaEval, 82.14$\to$81.41 on MT-Bench) while showing marginal improvement on pointwise \textsc{BiGGen} (0.434$\to$0.441 Spearman).
This suggests that pairwise and pointwise rubrics require different properties---pairwise rubrics must highlight discriminative dimensions, while pointwise rubrics need calibrated scoring anchors---and mixing both objectives dilutes the training signal for each.

\paragraph{Alternative Meta-Judge}
Our main experiments use Claude Sonnet 4 as the meta-judge $\mathcal{M}$ for collecting rubric preferences.
We ablate this choice by substituting Llama 3.1 70B as the meta-judge.
The weaker meta-judge substantially degrades AlpacaEval performance (70.09$\to$67.50) while slightly improving MT-Bench (82.14$\to$82.55) and \textsc{BiGGen} (74.25$\to$74.46).
Overall, using a stronger meta-judge provides more reliable reward signals, particularly on benchmarks where subtle quality differences matter (e.g., AlpacaEval's open-ended instructions).

\paragraph{Generator Model Size}
We additionally ablate the rubric generator model size by fine-tuning Llama 3.1 8B (vs.\ Qwen3 14B in the main paper).
Table~\ref{table:ablation_generator_size} reports results with the fine-tuned 8B generator evaluated across three judge models.
The 8B generator consistently underperforms the Qwen3 14B generator (Table~\ref{table:preference_learning}), which we attribute to the smaller model's limited capacity for generating diverse, high-quality rubric candidates during reward signal collection.

\begin{table*}[h]
  \centering
  \small
  \begin{tabular}{l rrr rr }
\toprule
& \multicolumn{3}{c}{Pairwise}  & \multicolumn{2}{c}{Pointwise} \\
\cmidrule(lr){2-4} \cmidrule(lr){5-6}
& AlpacaEval & MT-Bench &  \textsc{BiGGen} &  HelpSteer2  & \textsc{BiGGen}  \\

\midrule
Fine-tuned Llama 3.1 8B Generator\\

~~~Judge: Llama 3.1 8B \\
~~~~~~Iteration 1 (pairwise)                       & 68.22         & 73.22         & 67.19 & ---         & --- \\
~~~~~~Iteration 1 (pointwise)                      & ---   & ---   & ---                           & 0.200/0.268 & 0.311/0.321 \\
\addlinespace
~~~Judge: Llama 3.1 70B \\
~~~~~~Iteration 1 (pairwise)                       & 68.54         & 80.98         & 71.32 & ---         & --- \\
~~~~~~Iteration 1 (pointwise)                      & ---   & ---   & ---   & 0.270/0.342 & 0.380/0.392 \\
\addlinespace
~~~Judge: Claude Sonnet 4 \\
~~~~~~Iteration 1 (pairwise)                       & 70.09         & 82.14         & 74.25 & ---         & --- \\
~~~~~~Iteration 1 (pointwise)                      & ---   & ---   & ---   & 0.392/0.428 & 0.434/0.441 \\

\bottomrule
\end{tabular}

  \caption{Results with fine-tuned Llama 3.1 8B as rubric generator (Iteration~1), evaluated across three judge models. Compared to the Qwen3 14B generator (Table~\ref{table:preference_learning}), the 8B generator yields lower performance across most benchmarks, suggesting that generator capacity is an important factor for rubric quality.}
  \label{table:ablation_generator_size}
\end{table*}

\section{Qualitative Analysis and Case Studies}
\label{sec:appendix-qualitative}

\subsection{Analysis of Rubric Framing Shift}

The key qualitative shift across iterations is from \emph{prescriptive} criteria to \emph{dimension-labeled} criteria.
Prescriptive criteria (e.g., ``The response clearly explains...'', ``The answer includes...'') describe what a good response should contain, functioning as binary checklists.
Dimension-labeled criteria (e.g., ``Accuracy of ...'', ``Clarity and coherence in ...'') frame evaluation around abstract axes along which responses can be compared.

We quantify this shift by counting criteria that begin with abstract dimension labels versus prescriptive ``The response...'' patterns.
For pairwise rubrics, prescriptive criteria drop from 47.2\% (base) to 39.5\% (Iteration~1) to 24.9\% (Iteration~2), while dimension-labeled criteria rise from 13.8\% to 13.1\% to 26.4\%.
The trend is even more pronounced for pointwise rubrics: prescriptive criteria decline from 38.6\% to 33.5\% to 0.9\%.

This reframing explains why preference fine-tuning improves pairwise evaluation more than pointwise.
Dimension-labeled criteria guide the judge to compare responses \emph{along specific axes}, providing clear discriminative signal for relative judgments.
Prescriptive criteria are more binary---either both responses satisfy the criterion or neither does---offering less signal for distinguishing quality differences between two responses.
For pointwise scoring, however, abstract dimensions without calibrated score-level anchors make consistent absolute scoring harder, explaining the smaller gains on pointwise benchmarks.

\subsection{Case Studies: Rubric Evolution Across Iterations}

We present representative examples showing how fine-tuning transforms rubric framing and improves judge accuracy.
All pairwise case studies use Claude Sonnet 4 as judge; the base rubric is generated by Claude Sonnet 4 itself (training-free), while the Iteration~1/2 rubrics are from the fine-tuned Qwen3 14B generator.
The pointwise case study uses Claude Sonnet 4 as judge with rubrics from the base Qwen3 14B generator vs.\ the fine-tuned Qwen3 14B (Iteration~2).

\begin{figure*}[t]
  \centering
  \begin{tcolorbox}[
    width=\linewidth,
    colback=white,
    colframe=gray!50,
    arc=1.5mm,
    boxrule=0.5pt,
    title={\textbf{\textsf{Case Study 1: Persuasive Writing (Pairwise --- Base incorrect, Iteration~2 correct)}}},
    fonttitle=\small,
    fontupper=\small\textsf{},
    bottom=4pt, top=4pt,
    left=4pt, right=4pt
  ]

  \textbf{Prompt:} Write a persuasive email to convince your introverted friend, who dislikes public speaking, to volunteer as a guest speaker at a local event. Use compelling arguments and address potential objections. Please be concise. \textit{[Follow-up:]} Can you rephrase your previous answer and incorporate a metaphor or simile in each sentence?

  \vspace{0.3em}
  \textit{Judge verdict: Human=Response~2, Base=Response~1 (\ding{55}), Iteration~2=Response~2 (\ding{51})}

  \vspace{0.3em}
  \textbf{Response~1} (follow-up turn only): ``My answer was like a magnet, attracting the attention of everyone in the room. My thoughts were like buckshot, flying in all directions at once. I tried to explain my position, but it was like trying to nail Jell-O to a wall. My friend's argument was like a snowball, growing in size and complexity the more I tried to ignore it. The conversation soon became like a game of ping pong, with each of us batting ideas back and forth. Finally, we reached a compromise that was like a ray of sunshine, bringing a much-needed sense of resolution.''

  \vspace{0.3em}
  \textbf{Response~2} (follow-up turn only, truncated): ``I'm writing to invite you to volunteer as a guest speaker at a local event for [Event Name]. Just like a bubbly champagne, you'll be bursting with confidence on stage, sparkling with excitement and enthusiasm. [...] I know you're an introvert and dislike public speaking, but you will be like a superhero, saving the day on stage. I can help you prepare for the event by sharing a few tips and tricks that will make you more comfortable and confident on stage. [...]''

  \tcblower

  \textbf{Base rubric} (Claude Sonnet 4, training-free)\textbf{:}
  \begin{compactitem}
      \item The response demonstrates understanding of introversion and public speaking anxiety by acknowledging these concerns with empathy and sensitivity.
      \item The email uses compelling, personalized arguments that highlight specific benefits or opportunities relevant to an introverted person's values and interests.
      \item The response proactively addresses common objections (fear, preparation time, audience size, etc.) with practical solutions or reassuring counterpoints.
      \item Each sentence effectively incorporates a relevant metaphor or simile that enhances the persuasive message rather than feeling forced or distracting.
      \item The email maintains an appropriate tone that balances encouragement with respect for the friend's comfort zone and autonomy.
  \end{compactitem}

  \vspace{0.5em}
  \textbf{Iteration~2 rubric} (Qwen3 14B, fine-tuned)\textbf{:}
  \begin{compactitem}
      \item The response effectively uses a metaphor or simile in each sentence to enhance persuasiveness and clarity.
      \item The email addresses the friend's introversion and fear of public speaking with empathy and logical, compelling arguments.
      \item Potential objections, such as anxiety or discomfort, are clearly acknowledged and countered with thoughtful reasoning.
      \item The tone is warm, supportive, and appropriately tailored to an introverted audience.
      \item The email remains concise while fully addressing the request and maintaining persuasive power.
  \end{compactitem}

  \end{tcolorbox}
  \caption{Multi-turn persuasive writing task (MT-Bench). Response~1 uses metaphors but abandons the persuasive email format entirely, while Response~2 maintains the email structure with metaphors in each sentence. The base rubric buries the key follow-up constraint (metaphor/simile in each sentence) as criterion \#4, while the Iteration~2 rubric elevates it to criterion \#1. This prioritization helps the judge correctly identify which response better satisfies the follow-up instruction.}
  \label{fig:case_study_persuasive}
\end{figure*}

\begin{figure*}[t]
  \centering
  \begin{tcolorbox}[
    width=\linewidth,
    colback=white,
    colframe=gray!50,
    arc=1.5mm,
    boxrule=0.5pt,
    title={\textbf{\textsf{Case Study 2: Family Relationships (Pairwise --- Base incorrect, Iteration~1/2 correct)}}},
    fonttitle=\small,
    fontupper=\small\textsf{},
    bottom=4pt, top=4pt,
    left=4pt, right=4pt
  ]

  \textbf{Prompt:} A is the father of B. B is the father of C. What is the relationship between A and C? \textit{[Follow-up:]} If C is the son of D, D is the father of E, E is the son of X, and X is the father of Y, and Y is the father of Z, what's the relationship between A and Z in terms of generations and also the familial relationship in words?

  \vspace{0.3em}
  \textit{Judge verdict: Human=Response~1, Base=Response~2 (\ding{55}), Iteration~1=Response~1 (\ding{51}), Iteration~2=Response~1 (\ding{51})}

  \vspace{0.3em}
  \textbf{Response~1:} Turn~1: ``A and C are grandfather and grandson.'' Turn~2: ``A and Z are great-grandfather and great-grandson.''

  \vspace{0.3em}
  \textbf{Response~2} (truncated): Turn~1: ``A is the grandfather of C.'' Turn~2: ``Let's break down the relationships step by step: 1.~A is the father of B [...] Now, we can determine the relationship between A and Z: Since A is the grandfather of C, and Z is in the same generation as C, A is the great-grandfather of Z. In terms of generations, A is two generations above Z.''

  \tcblower

  \textbf{Base rubric} (Claude Sonnet 4, training-free)\textbf{:}
  \begin{compactitem}
      \item Correctly identifies that A is the grandfather of C in the first part of the question.
      \item Accurately traces the complete family lineage from A through all intermediate generations to Z without errors.
      \item Provides the correct generational distance between A and Z (A is 6 generations older than Z).
      \item Uses precise familial terminology to describe the relationship (A is the great-great-great-great-grandfather of Z).
      \item Shows clear logical reasoning by systematically working through each parent-child relationship in the chain.
  \end{compactitem}

  \vspace{0.5em}
  \textbf{Iteration~2 rubric} (Qwen3 14B, fine-tuned)\textbf{:}
  \begin{compactitem}
      \item Accuracy in determining the direct familial relationship between A and C based on the given information.
      \item Correct identification of the generational gap between A and Z, tracing the lineage through all intermediate individuals.
      \item Clear and precise description of the familial relationship between A and Z using appropriate terminology.
      \item Logical consistency in connecting each person in the lineage chain without introducing errors or assumptions.
      \item Conciseness and clarity in presenting both the generational difference and the familial relationship in a single, well-structured response.
  \end{compactitem}

  \end{tcolorbox}
  \caption{Multi-turn reasoning task (MT-Bench) requiring family relationship inference. The base rubric \emph{bakes in specific answers}: criterion \#3 states ``A is 6 generations older than Z'' and criterion \#4 specifies ``great-great-great-great-grandfather,'' which may not be correct given the ambiguous lineage. This prescriptive framing biases the judge toward the verbose Response~2 that attempts detailed step-by-step reasoning. The Iteration~2 rubric avoids committing to specific answers, using neutral evaluation dimensions (``Accuracy in determining...'', ``Logical consistency...'') that let the judge assess correctness without being biased by the rubric itself.}
  \label{fig:case_study_reasoning}
\end{figure*}

\begin{figure*}[t]
  \centering
  \begin{tcolorbox}[
    width=\linewidth,
    colback=white,
    colframe=gray!50,
    arc=1.5mm,
    boxrule=0.5pt,
    title={\textbf{\textsf{Case Study 3: Tool Use (Pointwise --- Base score=2, Iteration~1/2 score=4, Human=5)}}},
    fonttitle=\small,
    fontupper=\small\textsf{},
    bottom=4pt, top=4pt,
    left=4pt, right=4pt
  ]

  \textbf{Prompt:} In this instance you will be assigned to use the following tools. 1.~\textbf{Flight Booking}: [...specify departure city, destination city, and travel dates...] 2.~\textbf{Hotel Reservation}: [...] 3.~\textbf{Activity Planner}: [...] Plan a trip from New York to Tokyo, including flight, hotel, and activities for 3 days.

  \vspace{0.3em}
  \textit{Judge scores: Human=5, Base=2, Iteration~1=4, Iteration~2=4}

  \vspace{0.3em}
  \textbf{Response} (truncated): ``Hi, I'm happy to help you with your question about planning a trip from New York to Tokyo. [...] 1.~First, let's use the Flight Booking tool: \texttt{<flight\_booking> Departure City: New York | Destination City: Tokyo | Departure Date: 2023-03-20 | Return Date: 2023-03-22 </flight\_booking>} [...] 2.~Next, let's use the Hotel Reservation tool: \texttt{<hotel\_reservation> [...] </hotel\_reservation>} [...] 3.~Finally, let's use the Activity Planner tool: \texttt{<activity\_planner> [...] Interests: Cultural Sites, Nature, Dining </activity\_planner>} [...] By using these tools, you can easily plan your trip from New York to Tokyo. Have a great trip!''

  \tcblower

  \textbf{Base rubric} (Qwen3 14B, training-free)\textbf{:}
  \begin{compactitem}
      \item The response correctly uses the specified format for the Flight Booking tool with accurate departure and destination cities and travel dates.
      \item The response correctly uses the specified format for the Hotel Reservation tool with accurate destination city, check-in and check-out dates, and appropriate preferences.
      \item The response correctly uses the specified format for the Activity Planner tool with accurate destination city, travel dates, and relevant interests.
      \item The response includes all three required tools (Flight Booking, Hotel Reservation, and Activity Planner) in the correct order and without missing any required parameters.
      \item The response does not include any extra text or deviation from the required format, strictly adhering to the tool invocation syntax.
  \end{compactitem}

  \vspace{0.5em}
  \textbf{Iteration~2 rubric} (Qwen3 14B, fine-tuned)\textbf{:}
  \begin{compactitem}
      \item \textbf{Correct Use of Tool Formats}: The response must accurately use the specified XML-like format for each tool (flight\_booking, hotel\_reservation, activity\_planner) with all required parameters filled in correctly.
      \item \textbf{Completeness of Information}: The response must include all necessary details for each tool, such as departure and destination cities, dates, preferences, and interests, without missing any required fields.
      \item \textbf{Logical Trip Planning}: The response should demonstrate a coherent and realistic trip plan, with flights, hotels, and activities that align in terms of dates and destination.
      \item \textbf{Adherence to Instructions}: The response must strictly follow the given instructions, including not attempting to generate fake results but only the correct tool calls.
      \item \textbf{Clarity and Structure}: The response should be well-organized, with each tool call clearly separated and formatted as specified, ensuring readability and proper execution by the external tools.
  \end{compactitem}

  \end{tcolorbox}
  \caption{Tool-use task (\textsc{BiGGen Bench}, pointwise evaluation) where trained rubrics produce more accurate scores. The base rubric is a purely format-checking checklist (``correctly uses the specified format...''), which penalizes any minor deviation regardless of overall response quality. The Iteration~2 rubric introduces higher-level evaluation dimensions such as ``Logical Trip Planning'' and ``Clarity and Structure'' that capture overall quality beyond syntactic correctness. This enables the judge to assign a score (4) closer to the human assessment (5) rather than the overly strict base score (2).}
  \label{fig:case_study_tool}
\end{figure*}

\end{document}